\documentclass[conference]{IEEEtran}
\usepackage{cite}
\usepackage{amsmath,amssymb,amsfonts}
\usepackage{graphicx}
\usepackage{textcomp}
\usepackage{xcolor}
\usepackage{url}
\usepackage{algorithm}
\usepackage{algpseudocode}
\usepackage{subcaption}
\UseRawInputEncoding

    \usepackage{pdflscape} % enable landscape mode
    \usepackage{multirow} % multirow
    \usepackage{longtable} % multipage
    \usepackage{array} % fixed width
    
    \usepackage{booktabs} % top/mid/bottomrule
    \usepackage{makecell} % tableception
    \usepackage{diagbox} % \diagbox{}{}
    \usepackage{rotating}
\def\BibTeX{{\rm B\kern-.05em{\sc i\kern-.025em b}\kern-.08em
    T\kern-.1667em\lower.7ex\hbox{E}\kern-.125emX}}
\begin{document}

\title{Gait-learning with morphologically evolving robots generated by L-system}
\author{\IEEEauthorblockN{1\textsuperscript{st} Jie Luo}
\IEEEauthorblockA{\textit{Artificial Intelligence dept.} \\
\textit{Vrije Universiteit Amsterdam}\\
Amsterdam, The Netherlands \\
j2.luo@vu.nl}
\and
\IEEEauthorblockN{2\textsuperscript{nd} Daan Zeeuwe}
\IEEEauthorblockA{\textit{Artificial Intelligence dept.} \\
\textit{Vrije Universiteit Amsterdam}\\
Amsterdam, The Netherlands \\
zeeuwed001@gmail.com}
\and
\IEEEauthorblockN{3\textsuperscript{rd} Agoston E. Eiben}
\IEEEauthorblockA{\textit{Artificial Intelligence dept.} \\
\textit{Vrije Universiteit Amsterdam}\\
Amsterdam, The Netherlands \\
a.e.eiben@vu.nl}
}

\maketitle

\begin{abstract}
When controllers (brains) and morphologies (bodies) of robots simultaneously evolve, this can lead to a problem, namely the brain \& body mismatch problem. In this research, we propose a solution of lifetime learning. We set up a system where modular robots can create offspring that inherit the bodies of parents by recombination and mutation. With regards to the brains of the offspring, we use two methods to create them. The first one entails solely evolution which means the brain of a robot child is inherited from its parents. The second approach is evolution plus learning which means the brain of a child is inherited as well, but additionally is developed by a learning algorithm - RevDEknn. We compare these two methods by running experiments in a simulator called Revolve and use efficiency, efficacy, and the morphology intelligence of the robots for the comparison. The experiments show that the evolution plus learning method does not only lead to a higher fitness level, but also to more morphologically evolving robots. This constitutes a quantitative demonstration that changes in the brain can induce changes in the body, leading to the concept of morphological intelligence, which is quantified by the learning delta, meaning the ability of a morphology to facilitate learning.
\end{abstract}


\section{Introduction}
In the field of Evolutionary Robotics, the majority of studies consider the evolution of brains with a fixed body. This is not surprising, considering that the joint evolution of morphologies and controllers implies two search spaces and the search space for the brain changes with every new robot body produced. Evolving morphologies and controllers of robots simultaneously leads to a problem which has been noted long ago, being the body-brain mismatch problem \cite{Eiben2013}: Even though parents have well-matching bodies and brains, recombination and mutation can shuffle the parental genotypes such that the resulting body and brain combination might not fit well. Consequently, causing sub-optimal behaviour in the offspring. The proposed solution is the addition of learning. As phrased in “If it evolves it needs to learn”.\cite{Eiben2020}

The main goal of this research is to investigate the effects of learning in morphologically evolving robots. Our Hypothesis is that: 1) with learning, the time for achieving the same fitness level is less than without. 2) the learning approach does not only lead to different fitness levels, but also to different robot morphologies. 

To this end, we set up a system where (simulated) modular robots can reproduce and create offspring that inherit the parents’ morphologies by crossover and mutation. Regarding the controllers, we implement two methods. The first one is with evolution only which means the brain of a robot child is inherited from its parents. The second approach is evolution plus learning which means the brain of a child is also inherited, but additionally, it is optimized by a learning algorithm. The comparison is based on three measures: efficiency, efficacy, and morphological intelligence.

\section{Related Work}

\subsection{Evolvable morphology}

The body \& brain mismatch issue has been noted long ago, several approaches have been proposed to mitigate this effect on the population.

Cheney \textit{et al.} \cite{Cheney2014} implemented a form of novelty protection in which `younger' robot designs were protected from individuals that survived for more generations. Protecting a novel individual will increase its chance to adapt the controller properly for its body. Novelty protection corresponds with implementing a single lifetime learning iteration every time a morphology is protected. Similarly, De Carlo \textit{et al.} \cite{de2020influences} implemented protection in the form of speciation within their NEAT algorithm. The preservation of diversity in the population allowed new morphologies to survive, thus reducing the effects of body-brain mismatch. 

Nygaard, Samuelsen, and Glette \cite{nygaard2017overcoming} demonstrated improvements in their ER system by introducing two phases during evolution. The first phase consists of both controller and morphology evolution, while during the second phase only the controller evolves in a fixed body. The results showed that, without the second phase, morphology and controller evolution led to sub-optimal controllers which required additional fine-tuning. 

In this paper, we use the Triangle of Life framework (Figure 1) to integrates evolution and life time learning \cite{Eiben2013}. The essence is to have newborn robots perform a learning process that optimizes their inherited brain quickly after birth. An important additional feature is that newborn robots are considered to be infertile (i.e., not eligible for reproduction) until they successfully finish the learning period. This prevents that inferior genetic information is propagated and thus it saves resources.

\begin{figure}
	\centering
	\includegraphics[width=80mm]{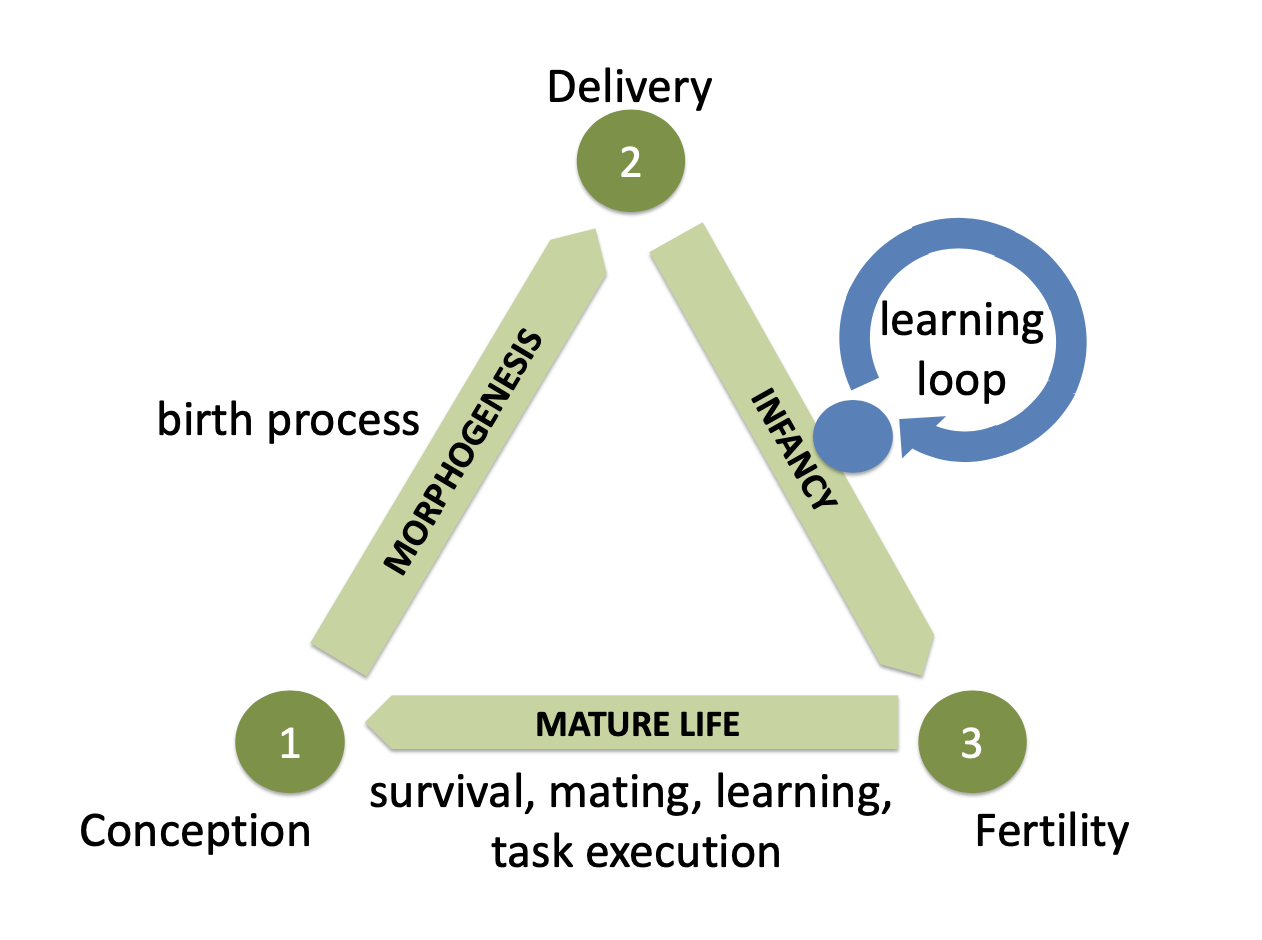}
	\caption{The life of triangle framework}
	\label{fig:life of triangle}
\end{figure}

\subsection{Controller learning algorithms}
Similar research has been done recently \cite{Diggelen} in which three learning algorithms have been compared whilst being applied to the CPG-based controllers to improve the weights. Namely: \textit{Evolutionary Strategies}, \textit{Bayesian Optimization} and \textit{Reversible Differential Evolution}(RevDE). The study shows that the shape of the fitness landscape in Evolutionary strategies hints to a possible bias for morphologies with many joints. This could be an unwanted property for the implementation of lifetime learning because we want an algorithm that can work consistently on different kinds of morphologies. Bayesian Optimization is good at sample efficiency, however it required much more time comparing to the other two methods due to the higher time-complexity. Therefore RevDE outperforms among these three algorithms. 

\textit{Differential Evolution} is a population-based Evolutionary algorithm (EA) that samples new candidates by perturbing the current population \cite{Storn1997}. The three main components in this method are as follows:

Differential mutation operator: A new candidate is generated by randomly picking a triplet from the population, $(x_i,x_j,x_k)\in X$, then $x_i$ is perturbed by adding a scaled difference between $x_j$ and $x_k$, that is:
        \begin{align}\label{eq:de1}
              y = x_i + F(x_j-x_k)
        \end{align}
where $F\in R_+$ is the scaling factor. 

Uniform crossover operator: the authors of \cite{Storn1997} proposed to sample a binary mask $m \in \{0, 1\}^D$ according to the Bernoulli distribution with probability p = P(md = 1) shared across all D dimensions, and calculate the final candidate according to the following formula:
        \begin{equation}\label{eq:de2}
              v = m \odot y+(1-m) \odot x_i
        \end{equation}

The last component is a selection mechanism: the authors use the “survival of the fittest” approach, i.e., combine the old population with the new one and select N candidates with the highest fitness values, i.e., the deterministic ($\mu$ + $\lambda$) selection.
%This variant of DE is referred to as “DE/rand/1/bin”, where rand stands for randomly selecting a base vector, 1 is for adding a single perturbation and bin denotes the uniform crossover. Sometimes it is called classic DE [25].

However, the mutation operator in DE perturbs candidates using other individuals in the population to generate a single new candidate. As a result, having too small a population could limit exploration of the search space and loose diversity. In order to overcome this issue, a modification of DE - \textit{Reversible DE} (RevDE) was proposed that utilized all three individuals to generate three new points in the following manner \cite{Tomczak2020}:

        \begin{equation}\label{eq:de3}
            \begin{split}
            y_1 &= x_i + F(x_j-x_k) \\
            y_2 &= x_j + F(x_k-y_1) \\
            y_3 &= x_k + F(y_1-y_2) 
            \end{split}
        \end{equation}
New candidates $y_1$ and $y_2$ could be further used to calculate perturbations using points outside the population. This approach does not follow a typical construction of an EA where only evaluated candidates are mutated. Further, we can express \eqref{eq:de3} as a linear transformation using matrix notation by introducing matrices as follows:

\begin{figure}[h]
\centering
  \includegraphics[width=70mm]{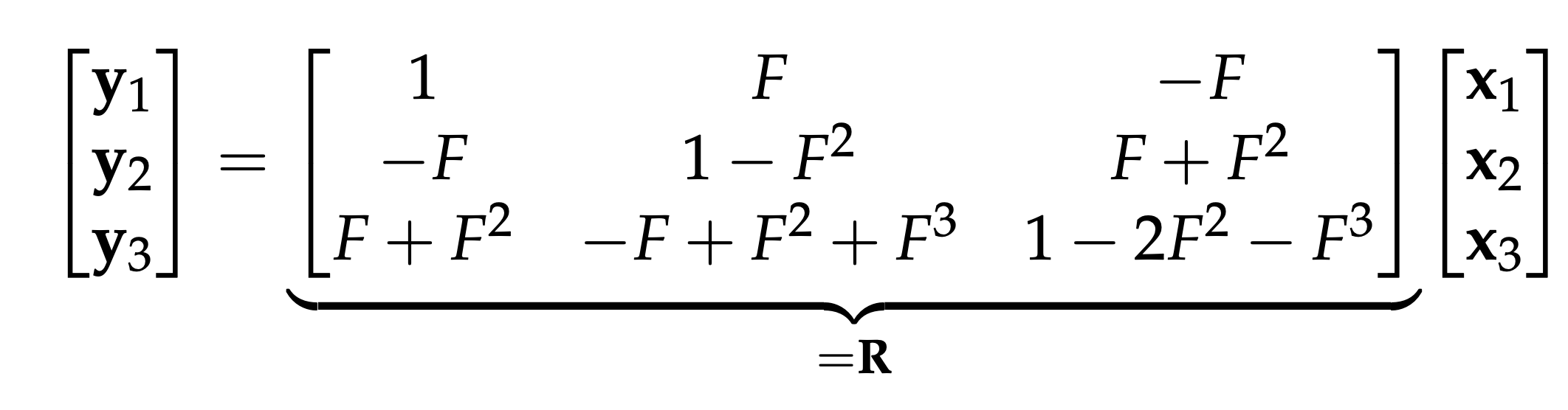}
  \label{fig:component}
\end{figure}

In order to obtain the matrix R, we need to plug $y_1$ to the second and third equation in (\ref{eq:de3}), and then $y_2$ to the last equation in (\ref{eq:de3}). As a result, we obtain M = 3N new candidate solutions and the linear transformation R is reversible.

However, generating 3N new candidates and evaluating all of them further comes with an extra computational cost while running the simulator. In this paper, we will use an advanced version of RevDE to alleviate this issue. This algorithm is introduced in the Algorithm section.

\section{Experiment Set-up}
The experiments have been carried out in Revolve (https://github.com/ci-group/revolve), a Gazebo based simulator which enables us to test the parts of the system as well as to set an entire environment for the complete evolutionary process. All experiments were performed using an infinite plane environment to avoid any extra complexity. We ran two experiments: experiment 1 works by running evolution alone. In this system, controllers are inheritable and the controller of the offspring is produced by applying crossover and mutation to the controllers of the parents. We refer to this experiment as Evolution Only throughout the paper. In experiment 2, controllers are not only evolvable, but also learnable. In this experiment, the controller of the offspring is produced by a learning algorithm that starts with the inherited brain. We refer to this experiment as Evolution + Learning throughout the paper.
%The experiments have been carried out in Revolve, a Gazebo-based simulation, in which the performance of the selected algorithm has been evaluated. The controller (brain) of each modular robot is learned / evolved independently in the simulation. Here we use our own custom simulator Revolve, based on Gazebo, that implements the components for running the Triangle of Life experiments \cite{Hupkes2018}. 
\subsection{Robot genotype (Body \& Brain)}
In this paper, we use a Lindenmayer-System (L-system) as the genetic representation \cite{Miras2020}. The grammar of a L-System is defined as a tuple G = (V, w, P ), where \\
– V, the alphabet, is a set of symbols containing replaceable and non-replaceable symbols. \\
– w, the axiom, is a symbol from which the system starts.\\
– P is a set of production-rules for the replaceable symbols.\\
The following didactic example illustrates the process of iterative-rewriting of an L-System. For a given number of iterations, each replaceable symbol is simultaneously replaced by the symbols of its production-rule. Given V = {X,Y,Z}, w = X and P = {X : {X,Y},Y : {Z},Z : {X,Z}}, the rewriting goes as follows:

\begin{center}
Iteration 0: X\\
Iteration 1: XY\\
Iteration 2: XY Z\\
Iteration 3: XY ZXZ\\
\end{center}

% after 3 iterations, this is the intermedia phenotype. mutation&crossover works on the rules/grammer

The construction of a phenotype (robot) from a genotype (grammar) is done by the following steps: 1. the axiom of the grammar is rewritten into a more complex string of symbols according to the production-rules of the grammar. 2. this string is decoded into a phenotype, one for the morphology (pointing to the current module) and one for the controller (pointing to the current sensor and the current oscillator). 

\subsection{Robot phenotype (Body)}
The robots in Revolve are based on the RoboGen framework \cite{Auerbach2014}.
We use a subset of 3D-printable components: one core component, one or more brick components, and one or more active hinges (see Figure \ref{fig:component}). Each robot's genotype describes its layout and consists of a tree-structure with the root node representing a core module from which further components branch out. Component types contain specific features described by its genotypical encoding dependant on a component’s type. These models are used in the simulation, but also could be used for 3D printing and construction of the real robots.

\begin{figure}[h]
  \includegraphics[width=\linewidth]{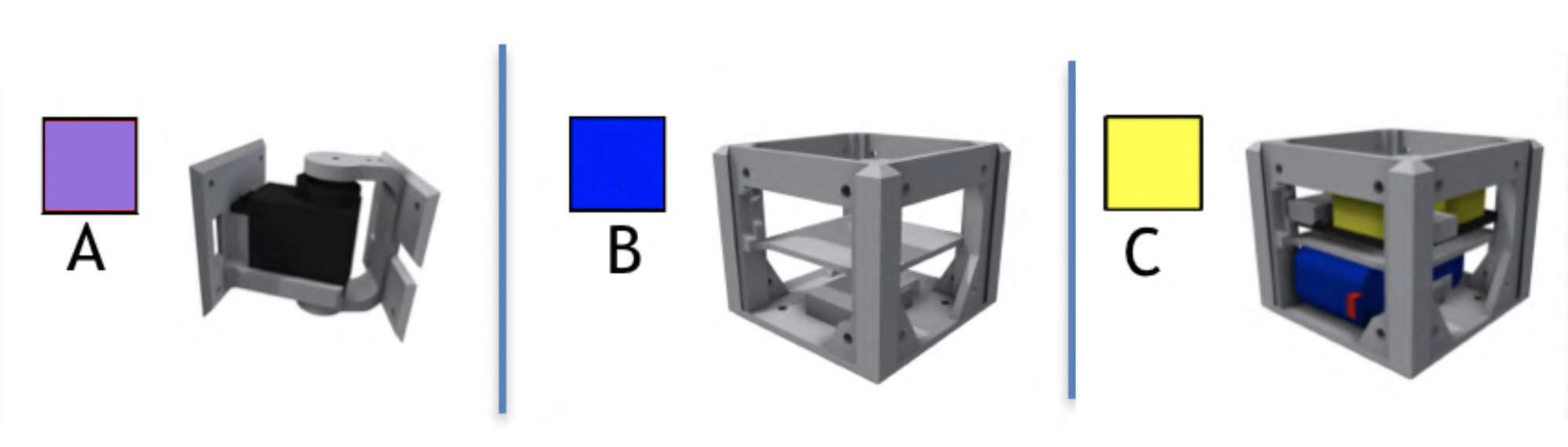}
  \caption{Modules of robots: The active hinge component (A) is a joint moved by a servomotor with attachment slots on both ends; The brick component (B) is a smaller cube with attachment slots on its lateral sides; The core component (C) which holds a controller board with battery is a large brick with four attachment slots on its lateral faces; }
  \label{fig:component}
\end{figure}

%As a test suite we chose the top four modular robots from Fuda's paper in order to compare the results using the RevDEknn algorithm. We refer to the four different size and shapes as gecko, insect, snake and Spider (see Figure \ref{fig:test_suite}).

\subsection{Robot phenotype (Brain)}
We use Central Pattern Generators (CPGs)-based controllers to drive the modular robots. CPGs are biological neural circuits that produce rhythmic outputs in the absence of rhythmic input. They are pairs of neurons ($x_i$,$y_i$) that drive rhythmic and stereotyped locomotion behaviors like walking, swimming, flying etc. in vertebrate species and they have been proven to perform well in modular robots \cite{Ijspeert2007}. 

\begin{figure}[hpt!]
\centering
  \includegraphics[width=70mm]{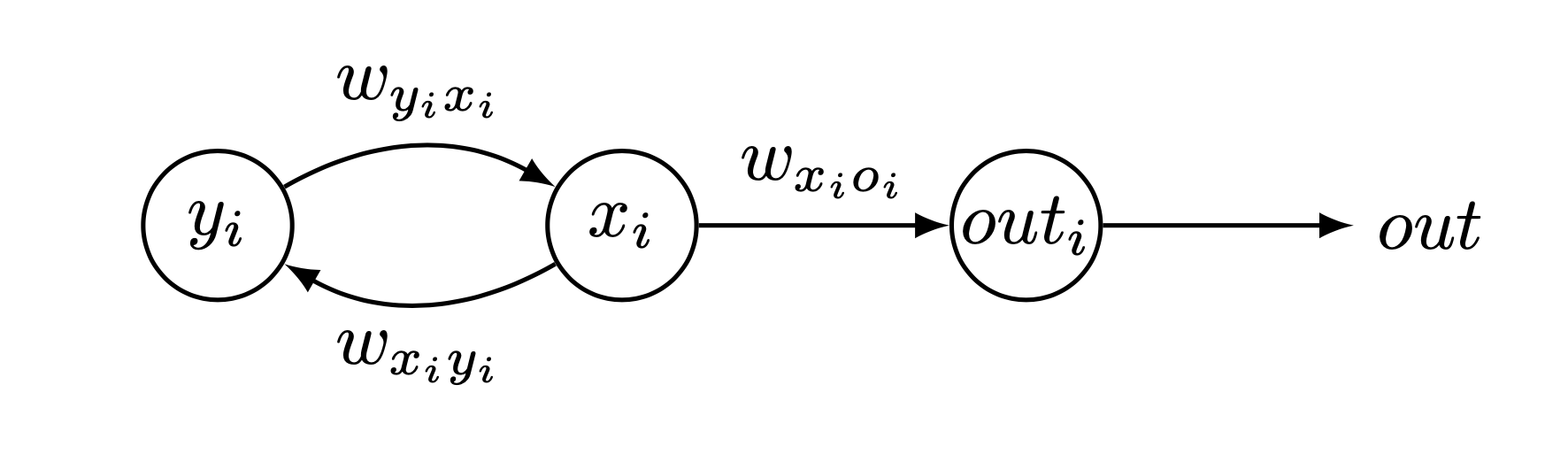}
  \caption{A single CPG. $i$ denotes the specific joint that is associated with this CPG. $w_{x_iy_i}$, $w_{y_ix_i}$, and $w_{x_io_i}$ denote the weights of the connections between the neurons, and out is the activation value of $out_i$-neuron that controls the servo in a joint $i$.}
  \label{fig:CPG_single}
\end{figure}

In this study, the controllers are optimized for gait-learning. Each robot joint is associated with a CPG that is defined by three neurons, an $x_i$-neuron, a $y_i$-neuron, and an $out_i$-neuron that are recursively connected as shown in Figure \ref{fig:CPG_single}. The change of a neuron's state is calculated by multiplying the activation value of the opposite neuron with a weight ($w$). So the $x_i$-neuron and $y_i$-neurons feed their activation values multiplied by weights $w_{x_iy_i}$ and $w_{y_ix_i}$ respectively to the $y_i$-neuron and $x_i$-neuron respectively. In our case, we use a variant of the sigmoid function, the hyperbolic tangent function (tanh), as the activation function of $out_i$-neurons to bound the output value in $[-1,1]$ due to the limited rotating angle of the joints.

\subsection{Algorithm}
It has been demonstrated that RevDE performs well to evolve controllers in modular robots for a given task \cite{Diggelen}.
However, it increases the computational cost of running the simulator by tripling the population. Here we introduce a surrogate model to overcome this issue. It uses the K-Nearest-Neighbor (K-NN) regressor to approximate the fitness values of the new candidates, then select N most promising points \cite{Weglarz-Tomczak2021}. We refer to this approach as RevDEknn.

The K-NN regression model is a non-parametric model that stores all previously seen individuals with their evaluations, and the prediction of a new candidate solution is an average over the K closest previously seen individuals (Table \ref{tab:RevDeknnparameters}). In this paper, we set K = 3.

The algorithm works as follows:\\
(1) initialize a population with X samples;\\
(2) evaluate the fitness of all X samples; \\
(3) perform a selection over the top samples to obtain vector $x_1$ in search space;\\
(4) randomly shuffle $x_1$ to create two additional vectors ($x_2,x_3$), and create the following new samples $y_1=x_1+F(x_2-x_3)$, $y_2=x_2+F(x_3-x_1)$ and $y_3=x_3+F(y_1-y_2)$. Apply uniform crossover with probability p between each $x_n$ and $y_n$; \\
(5) apply K-NN to predict the fitness value of new samples based on the 3 closest previously seen samples and repeat from (2). \\
(6) terminate when the maximum generation is reached.
 Following general recommendations in literature \cite{Pedersen2010} to obtain stable exploration/exploitation behaviour, the crossover probability p is fixed to a value of 0.9 and the scaling factor F is fixed to a value of 0.5. 

\begin{table}[ht!]
    \caption{Hyperparameters}
    %\setlength{\textfloatsep}{0.7\baselineskip plus 0.2\baselineskip minus 0.5\baselineskip}    
    %\small
    \label{tab:RevDeknnparameters}
    \centering

    \begin{tabular}{l ll}
    \hline
    \textbf{RevDEknn}         & Value     & Description \\
    \hline
    $X$             & 25        & Initial population size \\ 
    % $X$             & 25        & Initial Population size \\ 
    $x_1$           & 25        & Top samples size  \\
    $F$             & 0.5       & Scaling factor\\ 
    $p$             & 0.9       & Crossover probability \\ %CR
    $k$             & 3         & Number of Nearest-Neighbors \\ 
    $g$             & 10        & Number of Generations \\
     \hline

    \end{tabular} 
\end{table}

\begin{figure*}[hpt!]
  \includegraphics[width=\linewidth]{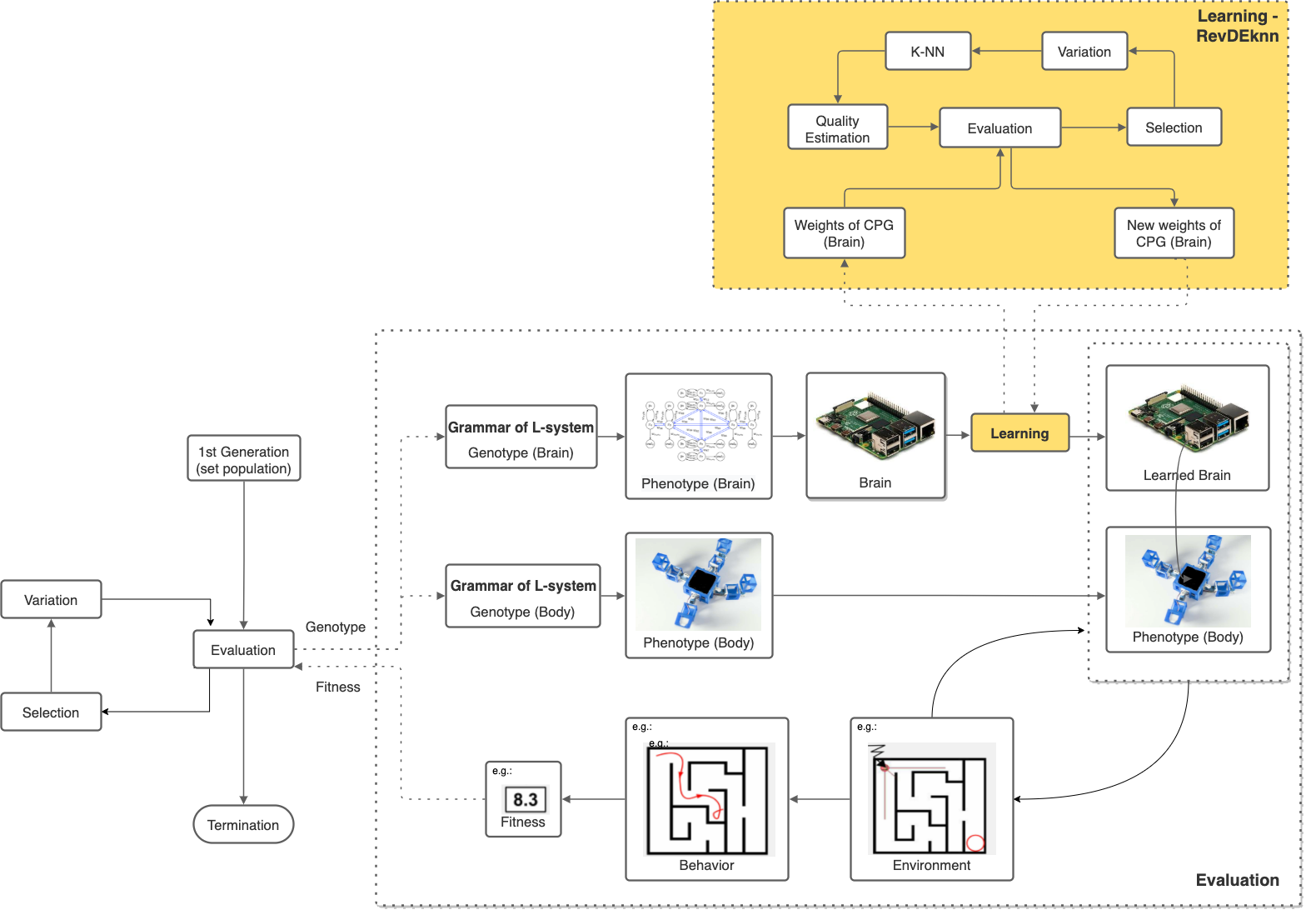}
  \caption{Evolution + Learning Framework: This is a general framework to make embodied robots via two interacting adaptive processes. An outer loop of evolution optimizes robot morphology via variation (mutation \& crossover) operations and an inner RevDEknn learning loop optimizes the parameters of a CPG controller (the yellow box). In the Evaluation box, we show examples of a robot morphology, controller, environment, behavior and fitness value.}
  \label{fig:E+L}
\end{figure*}

In this paper, we apply RevDEknn to change the weights of the CPGs of modular robots, for N CPGs, we have 3*N weights to improve their controllers for the task of gait-learning. For the whole big loop, we use Evolutionary Algorithm (EA). The whole process is illustrated in Figure \ref{fig:E+L}.

The code for carrying out the experiments is available online: \url{https://github.com/ci-group/revolve/tree/experiments/jlo_learning}. The pseudocode of combining EA and RevDEknn is shown below:

\begin{algorithm}[h!]
  \caption{EA+DevDEknn}
  \label{alg:hyperneat}
  \begin{algorithmic}[1]
    \State INITIALIZE robot population (genotypes + phenotypes with body and brain)  
    \State EVALUATE each robot  (evaluation delivers a fitness value)
    %\State EVALUATE each robot's CPG weights $\overrightarrow{\mathcal{W}}_1, \overrightarrow{\mathcal{W}}_2, ..., \overrightarrow{\mathcal{W}}_n$ to obtain the fitness $f_1, f_2, ...,f_n$.
    \While{not STOP-EVOLUTION}
    %\While{generation g<=30 \& g>1}
        \State SELECT parents; (based on fitness)
        \State RECOMBINE+MUTATE parents' bodies; (genotype)
        \State RECOMBINE+MUTATE parents' brains; (genotype)
        \State CREATE offspring robot body; (phenotype)
        \State CREATE offspring robot brain; (phenotype)
        
        \State INITIALIZE brain(s) for the learning process;
        \While{not STOP-LEARNING}
        %\While{learning l<=300 \& l>0}
            \State ASSESS offspring; (performance value)
            \State GENERATE new brain for offspring;
        \EndWhile 
        
        \State EVALUATE offspring w/ learned brain; (fitness value) 
        \State SELECT survivors / UPDATE population
          
    \EndWhile
    %\State \textbf{return} data $(\overrightarrow{\mathcal{W}}_{1:(g*n)},f_{1:(g*n)})$.
 \end{algorithmic}
\end{algorithm}
    
 \subsection{Experiment parameters}
An initial population of 50 robots is randomly generated in the first generation. In each generation 25 offspring are produced by selecting 25 pairs of parents through binary tournaments (with replacement) and creating one child per pair by crossover and mutation. From the top 25 parents plus 25 offspring, 50 individuals are selected for the next generation. The evolutionary process is terminated after 30 generations. In this research, the fitness value and performance value are the same. Therefore, for running Experiment Evolution Only, we perform $50+25*30=800$ fitness evaluations.

In Experiment Evolution + Learning, for each evolutionary process, we tested RevDEknn on gait learning during 750 (25 initial population * 3 by RevDE * 10 generations) learning trials with 250 assessments (750 divided by k=3 predictions) to simulate the robot’s limited field of view in the real-world. This resulted in $(50+25*30)*250=200,000$ fitness evaluations. 

We set the evaluation time to be 30 seconds to balance computing time and accurately evaluating a task as gait learning in which the fitness utilized was the speed (cm/s) of the robot’s displacement in any direction, notated as $s_x = (e_x-b_x)/ t$ where $b_x$ is x coordinate of the robot’s center of mass in the beginning of the simulation, $e_x$ is x coordinate of the robot’s center of mass at the end of the simulation, and t is the duration of the simulation. 

To sum up, for running these 2 experiments, we perform 200,800 evaluations which amounts to $200,800 * 30/60/60=1673.33$ hours of (simulated) time. In practice, it takes about 0.7 day to run these experiments on five computers with an Intel i7 CPU. All the experiments are repeated 10 times independently to get a robust assessment of the performance per data set. The experimental parameters we used in the experiments are described in Table \ref{tab:parameters}

\begin{table}[ht!]
\centering
\caption{Main experiment parameters}
\begin{tabular}{{p{0.25\linewidth} | p{0.1\linewidth}| p{0.5\linewidth}}}
\toprule
Parameters       & Value & Description                                    \\ \midrule
%Robots 			 & ~4     & Number of robot in \revolve (the test suite) 	\\
Population size  & ~50    & Number of individuals per generation     \\
Offspring size  & ~25    & Number of offspring produced per generation     \\
Mutation         & ~0.8   & Probability of mutation for individuals        \\ 
Crossover         & ~0.8   & Probability of crossover for individuals        \\ 
Generations      & ~30   & Termination condition for each run             \\ 
Learning trial  & ~750    & Number of the RevDEknn learning trials \\ 
Evaluation time  & ~30    & Duration of the test period per fitness evaluation in seconds \\ 
Tournament size  & ~2     & Number of individuals used in tournament selection 		 \\ 
%Gait Learning & ~5    & Number of test learning for each robot  \\
Repetitions      &  ~10    & Number of repetitions per experiment (each robot + gait-learning) \\ 
\bottomrule 
\end{tabular}
\label{tab:parameters}
\end{table}

\subsection{Performance measures}
To compare the two methods, we consider three performance indicators: \textit{efficiency}, \textit{efficacy}, and the \textit{morphologies intelligence}. 

% The assessments are based on inspecting the learning curves for each morphology. Additionally, to evaluate the overall performance we will aggregate the results over the morphologies ($N=20$) and perform statistical analysis (one-way ANOVA) to see if there is a significant difference between the learning methods (for $p\leq0.05$). An additional post-hoc analysis between the learners is done if we find a significant difference. Violations of the assumption on normality and/or equal variance result in taking a different appropriate statistical measure. The specific definitions of the performance indicators are as follows. 

\subsubsection{Efficiency \& Efficacy}

We measure efficacy by the quality achieved at the end of the evolutionary process. Since we consider gait learning here, the quality is defined by the speed of the robot. As this measure can be sensitive to `luck', we get more useful statistics by taking the average over 10 different runs. Thus here, the efficacy of a method is defined by the mean best fitness averaged over the 10 independent repetitions. Efficiency indicates how quickly the robot finds its best solution.

%\textit{Efficiency}: Efficiency indicates how quickly the robot finds its best solution. To this end, we use the Average number of Evaluations to Solution (AES) as defined in \cite{eib2003introduction}. More than indicating efficiency, AES can provide insight into the maximum number of evaluations required for our robots. If we overestimate the maximum number of evaluations required, then the AES values tend to be (much) lower compared to the evaluation limit set during the experiment.

\subsubsection{Morphological Descriptors}
For quantitatively assessing morphological traits of the robots, we utilized the following set of descriptors:

\textbf{Absolute Size}: Total number of modules of a robot body. It's a sum of all the structural bricks, hinges and one core-component with controller board.\\
\begin{center}
 absolute\_size = brick\_count + hinge\_count + $1$
 \end{center}

\textbf{Width}: The width of a body, excluding sensors.\\

\textbf{Proportion}: The length-width ratio of the rectangular envelope around the morphology. It is defined as following:
        \begin{align*}\label{eq:proportion}
              P =  \frac{p_s} {p_l}
        \end{align*}

where $p_s$ is the shortest side of the morphology, and $p_l$ is the longest side.

\textbf{Number of Bricks}: The number of structural bricks in the morphology.

\textbf{Relative Number of Limbs}: The number of extremities of a morphology relative to a practical limit. It is defined as following:

\begin{align*}
  L =
    \begin{cases}
      \frac{l}{l_{max}} & \text{if $l_{max} > 0$}\\
      0 & \text{otherwise}\\
    \end{cases}     
\end{align*}

\begin{align*}
  L_\text{max} =
    \begin{cases}
      \frac{2*(m-6)}{3} +(m-6)(mod 3) +4 & \text{if $l_{max} > 0$}\\
      m-1 & \text{otherwise}\\
    \end{cases}   
\end{align*}

where m is the total number of modules in the morphology, l the number of modules which have only one face attached to another module (except for the core-component) and $l_\text{max}$ is the maximum amount of modules with one face attached that a morphology with m modules could have, if containing the same amount of modules arranged in a different way.

\textbf{Number of Active Hinges}: Number of active hinges in the morphology. Activate hinge means a joint that has a motor, and non-activate is a passive joint (just the ‘bendable’ plastic).

We use the above six morphological descriptors to capture relevant robot morphological traits, and quantify the correlations between controller \& morphology search spaces.

\section{Experiment Results}
\subsection{Efficiency \& Efficacy}
\begin{figure}[h]
  \includegraphics[width=\linewidth]{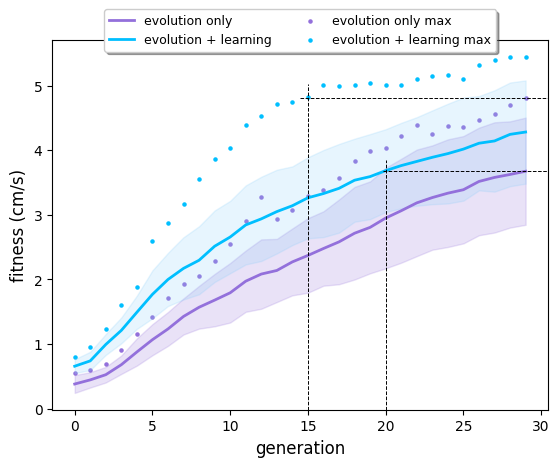}
  \caption{The blue and purple lines are the progression of the mean of fitness over 30 generations (avg. over 10 runs). The blue and purple dotted lines show the maximum fitness per generation. The black dotted lines mark generations, when the evo+learning method (after learning) achieved the levels of the fitness that the Evo only method managed to achieve only in the end of the evolutionary period.}
  \label{fig:fitness}
\end{figure}

\begin{figure}[h]
  \includegraphics[width=\linewidth]{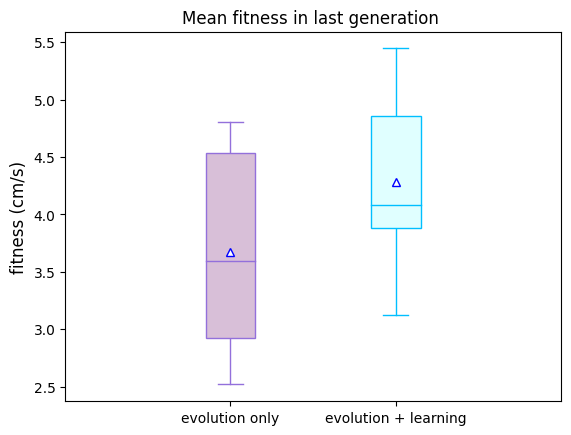}
  \caption{Comparison of fitness in the final generations. The blue triangles show the mean of the fitness.Significance levels for the Wilcoxon tests in the boxplot non-significant.}
  \label{fig:fitness_last_gen}
\end{figure}

Adding a life-time learning capacity to the system increased the speed of the robots, as depicted by Figures \ref{fig:fitness} and \ref{fig:fitness_last_gen}. This was expected for two reasons: a) the number of evaluations performed by Evolution + Learning is around 100 times higher than by Evolution Only; b) in Evolution + Learning, robots have time to fine-tune their controllers to the morphologies they were born with. Instead, we are interested in observing which method presents a faster growth of the average speed. In Figure \ref{fig:fitness}, the black line shows that around generation 20, Evolution + Learning had already obtained an average speed that took Evolution Only the whole evolutionary period to achieve, i.e., 30 generations. In another word, Evolution + Learning at generation 20 created only 550 robots and spent 55,000 evaluations while Evolution Only created 800 robots and 800 evaluations at generation 30. If we consider real physical robots, and assuming that the production cost of each robot (around 4 hours) is substantially higher than the evaluation cost (around 30 seconds), we can clearly see the advantage of introducing learning.

\begin{figure}[h]
  \includegraphics[width=\linewidth]{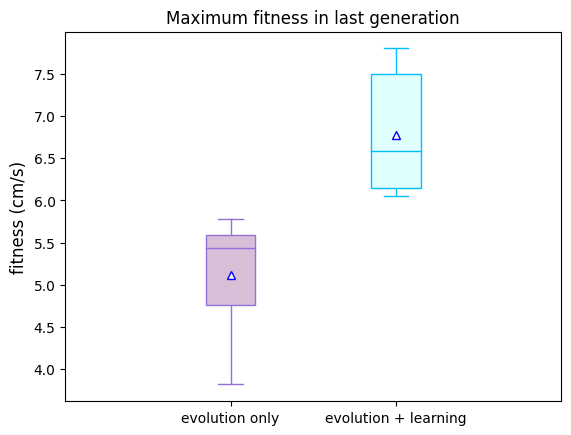}
  \caption{Comparison of maximum fitness in the final generations. The blue triangles show the mean of the fitness. Significance levels for the Wilcoxon tests in the boxplot are non-significant.}
  \label{fig:fitness_best_run}
\end{figure}

\subsection{Morphological Intelligence}
\begin{figure}[h]
  \includegraphics[width=\linewidth]{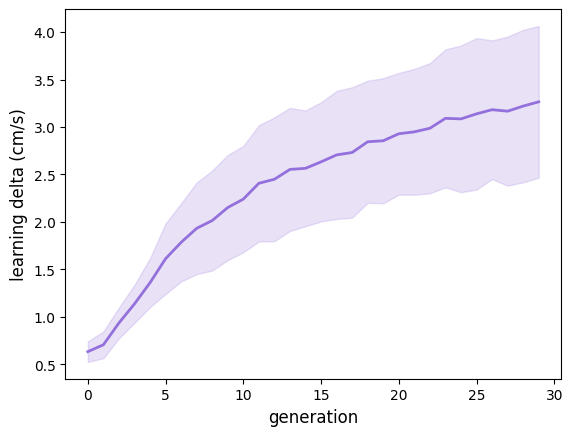}
  \caption{Learning $\Delta$: average speed after the parameters were learned minus average speed before the parameters were learned. Progression of the mean over 10 runs of the population.}
  \label{fig:learning delta}
\end{figure}

In this paper, we consider a new concept - Morphological Intelligence. Morphology influences how the brain learns. Some bodies are more suitable for the brains to learn with than others. How well the brain learns can be empowered by a better body. Therefore we define the intelligence of a body as a measure of how well it facilitates the brain to learn and achieve tasks, in this case, gait learning. In this paper, we quantify morphological intelligence by the learning delta, being speed after the parameters were learned minus speed before the parameters were learned. See Figure \ref{fig:learning delta}, where we see that the average learning $\Delta$ of the method Evolution + Learning, grows across the generations. 

This growth is very steady. The observation indicates that the life-time learning led the evolutionary search to more quickly exploit the high performing morphological properties. In other words, it was faster for the population to turn into morphologies that are big, disproportional, with few, long limbs which fit the brains better.

\subsection{Morphological Descriptors}
\begin{figure*}[hpt!]
    \begin{minipage}[c]{0.99\textwidth}
        \begin{minipage}[t]{0.3\textwidth}
            \centering
            \subfloat[]{\includegraphics[width=\textwidth]{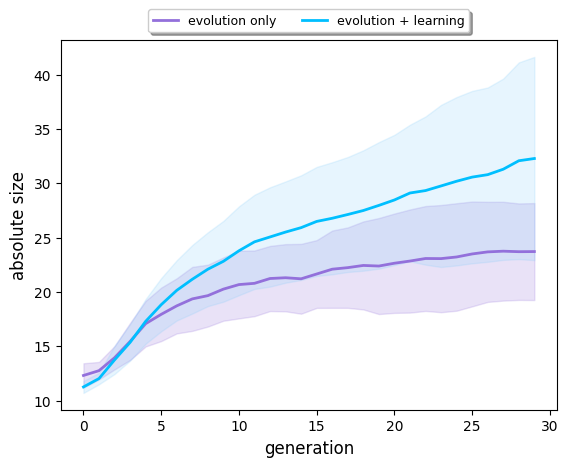}}
        \end{minipage}
        \hfill
        \begin{minipage}[t]{0.3\textwidth}
            \centering
            \subfloat[]{\includegraphics[width=\textwidth]{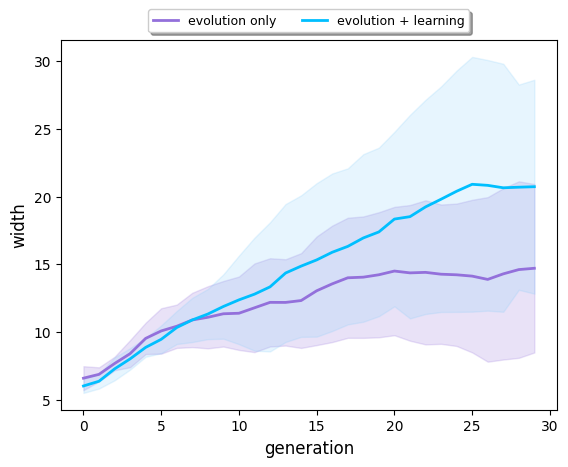}}
        \end{minipage}
        \hfill
        \begin{minipage}[t]{0.3\textwidth}
            \centering
            \subfloat[]{\includegraphics[width=\textwidth]{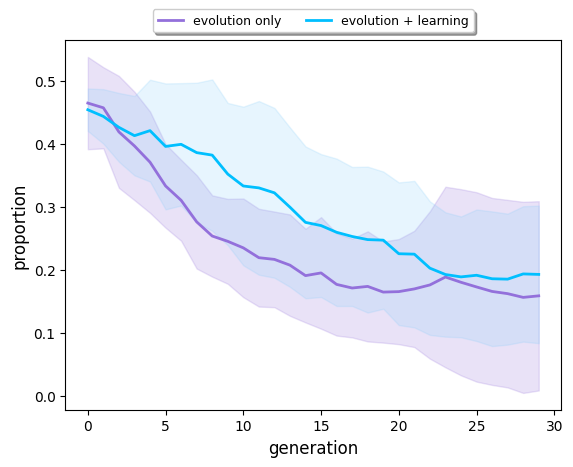}}
        \end{minipage}
    \end{minipage}
    
    \begin{minipage}[c]{0.99\textwidth}
        \begin{minipage}[t]{0.3\textwidth}
            \centering
            \subfloat[]{\includegraphics[width=\textwidth]{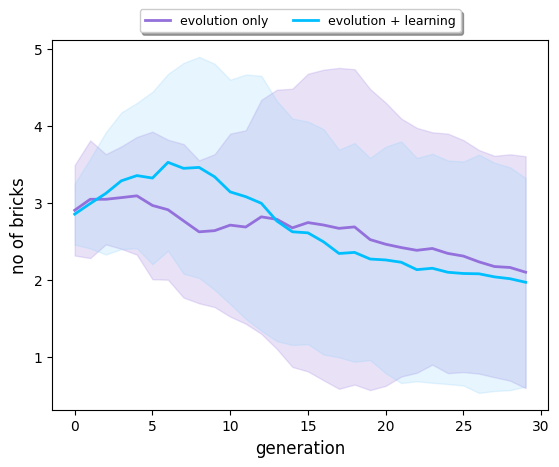}}
        \end{minipage}
        \hfill
        \begin{minipage}[t]{0.3\textwidth}
            \centering
            \subfloat[]{\includegraphics[width=\textwidth]{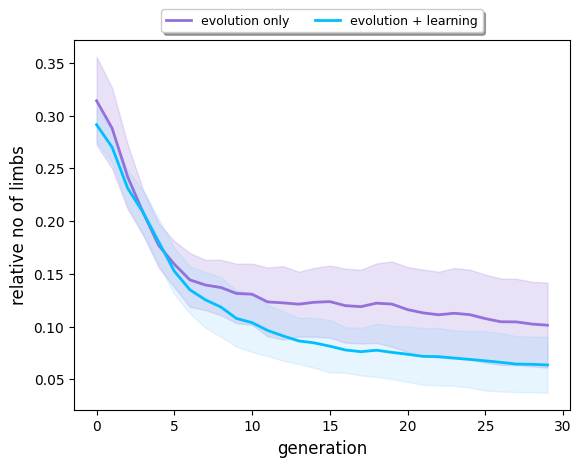}}
        \end{minipage}
        \hfill
        \begin{minipage}[t]{0.3\textwidth}
            \centering
            \subfloat[]{\includegraphics[width=\textwidth]{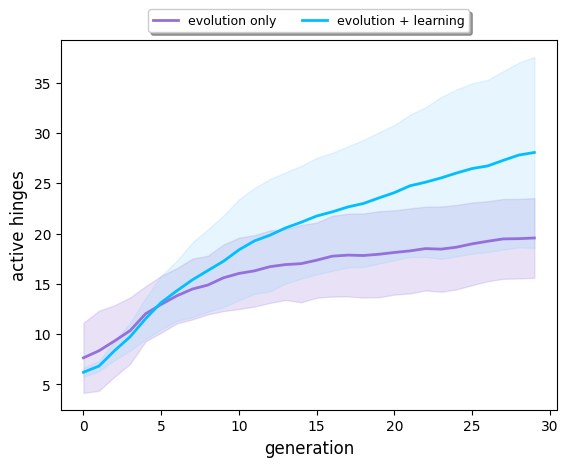}}
        \end{minipage}
    \end{minipage}
\vspace{-.5em}
\caption{We selected 6 morphological traits which give a clear trend over generations. Progression of the mean of different morphological descriptors averaged over 10 runs for the entire population. Shaded region denotes 95\% bootstrapped confidence interval. Evolution + learning robots tend to be bigger (a) along the evolution and wider in width (b) compared to robots evolved in Evolution only. Proportion (c) is the ratio of width and length of the robot morphology. Robots from both methods tend to have less number of bricks (d) \& limbs (e) and more active hinges (f) over generations.}
\label{fig:6 traits}
\end{figure*}

\begin{figure*}[hpt!]
    \begin{minipage}[c]{0.99\textwidth}
        \begin{minipage}[t]{0.47\textwidth}
            \centering
            \subfloat[]{\includegraphics[width=\textwidth]{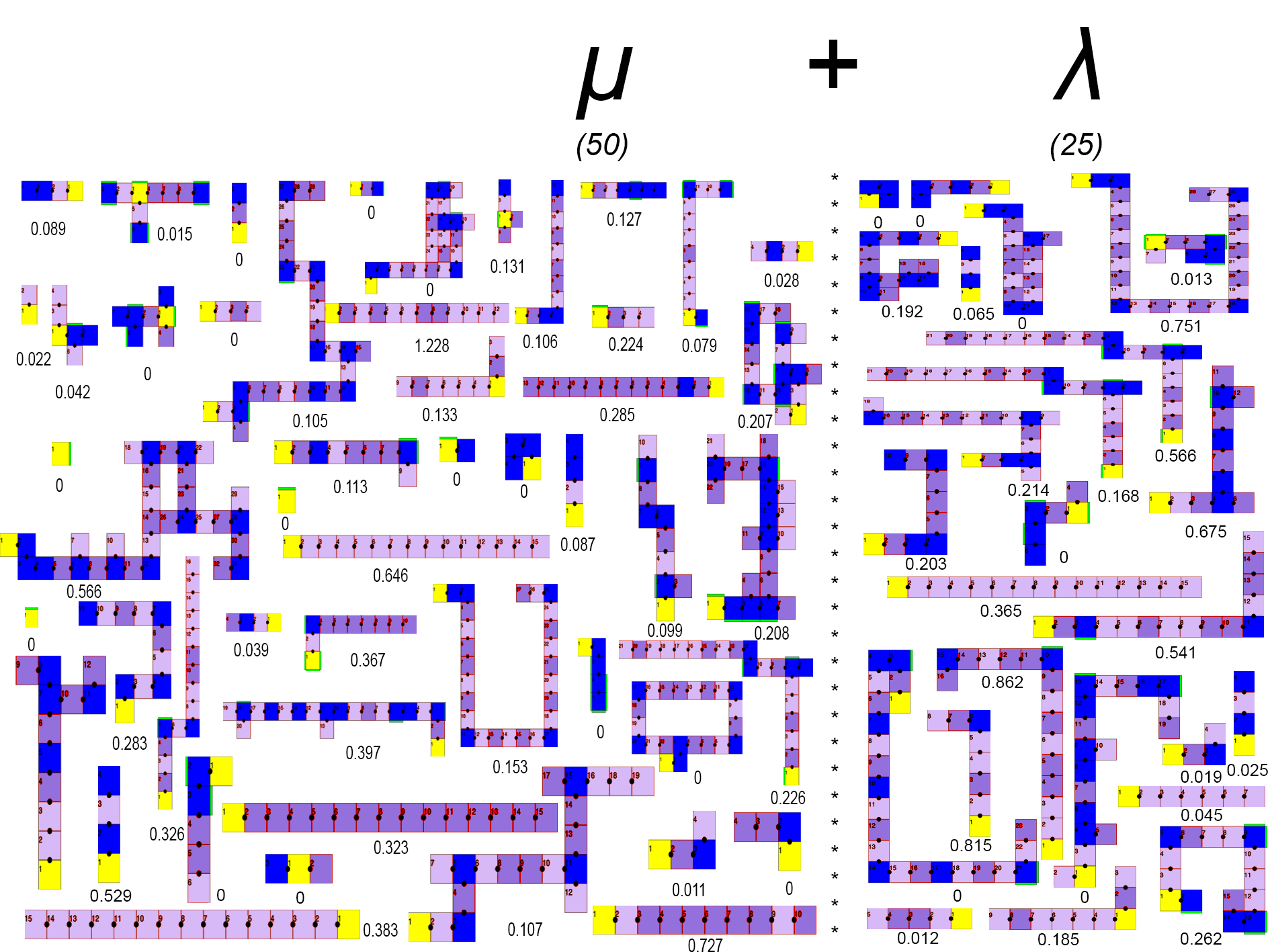}}
        \end{minipage}
        \hfill
        \begin{minipage}[t]{0.47\textwidth}
            \centering
            \subfloat[]{\includegraphics[width=\textwidth]{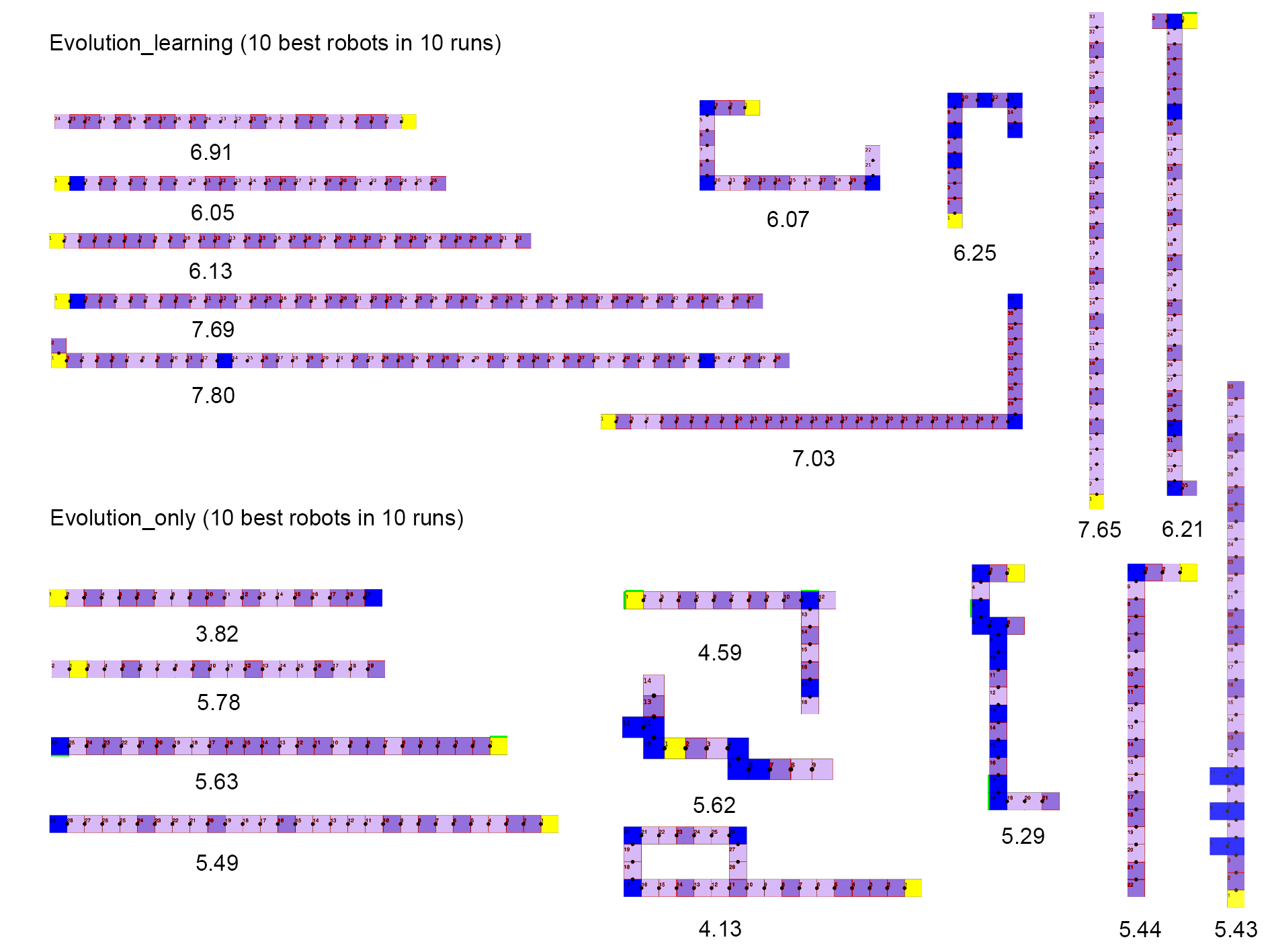}}
        \end{minipage}
    \end{minipage}
\caption{(a), an example of the initial randomly generated 50 morphologies in a run, using ($\mu+\lambda$)selection, 25 new off-springs were created. (b), the morphologies of best robots of each run for both control methods with the fitness value.}
\label{fig:morphology}
\end{figure*}

\begin{figure*}[hpt!]
    \centering
	\includegraphics[width=160mm]{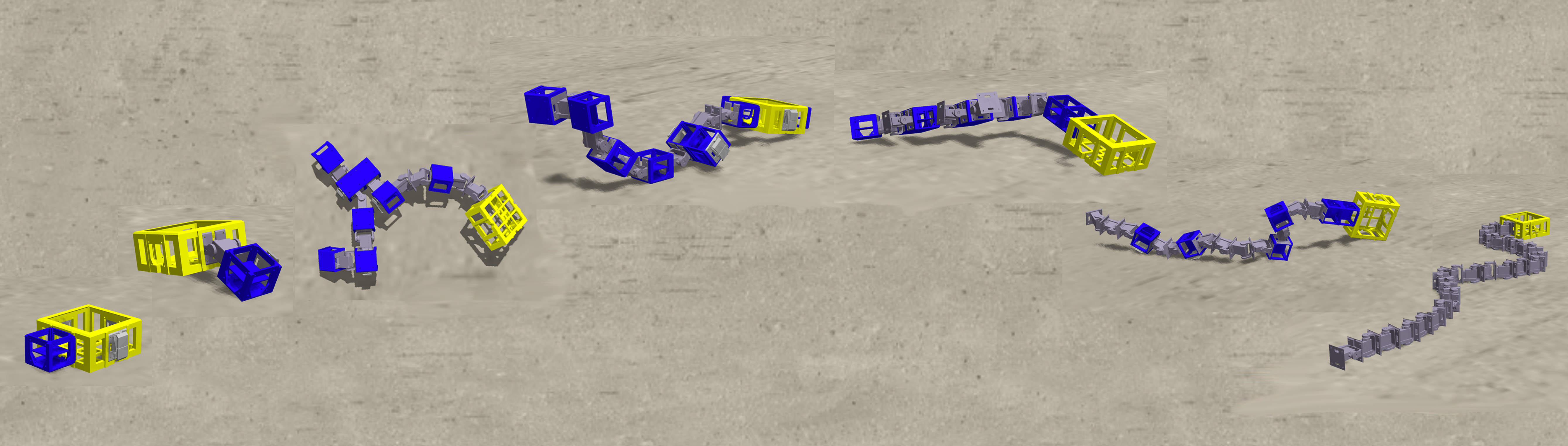}
	\caption {Learning/evolving robots in a plain terrain. Morphologies changes over generations}
	\label{fig:robot_evolve}
\end{figure*}

In \cite{Miras2020a}, a study utilizing this same robot framework observed a strong selection pressure for robots with few limbs, most often one single, long limb, i.e., a snake-like morphology. Furthermore, they demonstrated that by explicitly adding a penalty to having this morphological property, the population did indeed develop multiple limbs, nevertheless, these robots were much slower than the single-limb ones.
We  selected  6  morphological  traits  which  display  a  clear  trend  over  generations. In figure \ref{fig:6 traits}, we see the progression  of  the  mean of  different morphological descriptors averaged over 10 runs for the entire population. We observed a strong selection pressure for robots with few limbs, most often one single long limb, i.e., a snake-like morphology. Evolution + learning robots tend to be bigger along the evolution and wider in width compared to robots evolved in Evolution only. Robots from both methods tend to have fewer number of bricks, limbs and more active hinges over generations. We also observe that the generations evolve from having multiple limbs to having a single-limb, leading to higher speed. (Figure \ref{fig:morphology} and \ref{fig:robot_evolve}). A video showing examples of robots from both types of experiments can be found in https://www.youtube.com/watch?v=bH4A6sm4Umw.

\section{Conclusions and Future work}
Firstly, if we measure time by the number of generations, learning boosts evolvability in terms of efficiency as well as efficacy, i.e., solution quality at termination, once its growth curve was steeper and ended higher than that of the Evolution Only method. Of course, this is not a surprise, since the learning version performs much more search steps. However, a learning trial (testing another controller) is much cheaper than an evolutionary trial (making another robot), so we can firmly conclude the advantage of adding lifetime-learning to an evolutionary robot system.

Secondly, we have witnessed a change in the evolved morphologies when life-time learning was applied. In this paper, we introduced a concept - Morphological Intelligence, and quantified it as the learning delta. The results show how the brain can shape the body and thus affects task performance which in turn changes the fitness values that define selection probabilities during evolution.

For future work, we will work on Lamarckian evolution which use the inherited brain as starting point and pass the learned traits to the next generation. It means the genotype of the brain of the population will be changed in the evolution process. In order to make this happen, we have to carefully think about a suitable genetic representation that allows us to change the genetic traits easily. Moreover, we will implement more complex tasks than just gait learning. 

\bibliographystyle{IEEEtran}
\bibliography{library.bib}

\begin{thebibliography}{10}
\providecommand{\url}[1]{#1}
\csname url@samestyle\endcsname
\providecommand{\newblock}{\relax}
\providecommand{\bibinfo}[2]{#2}
\providecommand{\BIBentrySTDinterwordspacing}{\spaceskip=0pt\relax}
\providecommand{\BIBentryALTinterwordstretchfactor}{4}
\providecommand{\BIBentryALTinterwordspacing}{\spaceskip=\fontdimen2\font plus
\BIBentryALTinterwordstretchfactor\fontdimen3\font minus
  \fontdimen4\font\relax}
\providecommand{\BIBforeignlanguage}[2]{{%
\expandafter\ifx\csname l@#1\endcsname\relax
\typeout{** WARNING: IEEEtran.bst: No hyphenation pattern has been}%
\typeout{** loaded for the language `#1'. Using the pattern for}%
\typeout{** the default language instead.}%
\else
\language=\csname l@#1\endcsname
\fi
#2}}
\providecommand{\BIBdecl}{\relax}
\BIBdecl

\bibitem{Eiben2013}
A.~Eiben, N.~Bredeche, M.~Hoogendoorn, J.~Stradner, J.~Timmis, A.~Tyrrell, and
  A.~Winfield, ``{The Triangle of Life: Evolving Robots in Real-time and
  Real-space},'' no. September, pp. 1056--1063, 2013.

\bibitem{Eiben2020}
A.~E. Eiben and E.~Hart, ``{If it evolves it needs to learn},'' \emph{GECCO
  2020 Companion - Proceedings of the 2020 Genetic and Evolutionary Computation
  Conference Companion}, pp. 1383--1384, 2020.

\bibitem{Cheney2014}
N.~Cheney, J.~Clune, and H.~Lipson, ``{Evolved electrophysiological soft
  robots},'' \emph{Artificial Life 14 - Proceedings of the 14th International
  Conference on the Synthesis and Simulation of Living Systems, ALIFE 2014},
  no. 1994, pp. 222--229, 2014.

\bibitem{de2020influences}
M.~De~Carlo, D.~Zeeuwe, E.~Ferrante, G.~Meynen, J.~Ellers, and A.~Eiben,
  ``Influences of artificial speciation on morphological robot evolution,'' in
  \emph{2020 IEEE Symposium Series on Computational Intelligence (SSCI)},
  IEEE.\hskip 1em plus 0.5em minus 0.4em\relax Washington, DC, USA: IEEE
  Computer Society, 2020, pp. 2272--2279.

\bibitem{nygaard2017overcoming}
T.~F. Nygaard, E.~Samuelsen, and K.~Glette, ``Overcoming initial convergence in
  multi-objective evolution of robot control and morphology using a two-phase
  approach,'' in \emph{European Conference on the Applications of Evolutionary
  Computation}, Springer.\hskip 1em plus 0.5em minus 0.4em\relax New York, NY:
  Springer-Verlag, 2017, pp. 825--836.

\bibitem{Diggelen}
F.~V. Diggelen, ``{Comparing lifetime learning methods for morphologically
  evolving robots},'' in \emph{ACM}, no.~1, 2021.

\bibitem{Storn1997}
R.~M. Storn, ``{Differential evolution—A simple and efficient heuristic for
  global optimization over continuous spaces},'' \emph{Journal of Global
  Optimization}, pp. 131--141, 1997.

\bibitem{Tomczak2020}
J.~M. Tomczak, E.~Wȩglarz-Tomczak, and A.~E. Eiben, ``{Differential Evolution
  with Reversible Linear Transformations},'' \emph{GECCO 2020 Companion -
  Proceedings of the 2020 Genetic and Evolutionary Computation Conference
  Companion}, no.~1, pp. 205--206, 2020.

\bibitem{Miras2020}
K.~Miras, M.~{De Carlo}, S.~Akhatou, and A.~E. Eiben, ``{Evolving-Controllers
  Versus Learning-Controllers for Morphologically Evolvable Robots},''
  \emph{Lecture Notes in Computer Science (including subseries Lecture Notes in
  Artificial Intelligence and Lecture Notes in Bioinformatics)}, vol. 12104
  LNCS, no. April, pp. 86--99, 2020.

\bibitem{Auerbach2014}
J.~E. Auerbach, D.~Aydin, A.~Maesani, P.~M. Kornatowski, T.~Cieslewski,
  G.~Heitz, P.~R. Fernando, I.~Loshchilov, L.~Daler, and D.~Floreano,
  ``{Robogen: Robot generation through artificial evolution},''
  \emph{Artificial Life 14 - Proceedings of the 14th International Conference
  on the Synthesis and Simulation of Living Systems, ALIFE 2014}, pp. 136--137,
  2014.

\bibitem{Ijspeert2007}
A.~J. Ijspeert, A.~Crespi, D.~Ryczko, and J.~M. Cabelguen, ``{From swimming to
  walking with a salamander robot driven by a spinal cord model},''
  \emph{Science}, vol. 315, no. 5817, pp. 1416--1420, 2007.

\bibitem{Weglarz-Tomczak2021}
E.~Weglarz-Tomczak, J.~M. Tomczak, A.~E. Eiben, and S.~Brul,
  ``{Population-based parameter identification for dynamical models of
  biological networks with an application to Saccharomyces cerevisiae},''
  \emph{Processes}, vol.~9, no.~1, pp. 1--14, 2021.

\bibitem{Pedersen2010}
\BIBentryALTinterwordspacing
M.~Pedersen, ``{Good Parameters for Differential Evolution},''
  \emph{Evolution}, pp. 1--10, 2010. [Online]. Available:
  \url{http://www.hvass-labs.org/people/magnus/publications/pedersen10good-de.pdf}
\BIBentrySTDinterwordspacing

\bibitem{Miras2020a}
K.~Miras, E.~Haasdijk, K.~Glette, and A.~E. Eiben, ``{Effects of selection
  preferences on evolved robot morphologies and behaviors},'' \emph{ALIFE 2018
  - 2018 Conference on Artificial Life: Beyond AI}, no. April 2019, pp.
  224--231, 2020.

\end{thebibliography}

%\begin{thebibliography}{00}
% \bibitem{b1} G. Eason, B. Noble, and I. N. Sneddon, ``On certain integrals of Lipschitz-Hankel type involving products of Bessel functions,'' Phil. Trans. Roy. Soc. London, vol. A247, pp. 529--551, April 1955.
% \bibitem{b2} J. Clerk Maxwell, A Treatise on Electricity and Magnetism, 3rd ed., vol. 2. Oxford: Clarendon, 1892, pp.68--73.
% \bibitem{b3} I. S. Jacobs and C. P. Bean, ``Fine particles, thin films and exchange anisotropy,'' in Magnetism, vol. III, G. T. Rado and H. Suhl, Eds. New York: Academic, 1963, pp. 271--350.
% \bibitem{b4} K. Elissa, ``Title of paper if known,'' unpublished.
% \bibitem{b5} R. Nicole, ``Title of paper with only first word capitalized,'' J. Name Stand. Abbrev., in press.
% \bibitem{b6} Y. Yorozu, M. Hirano, K. Oka, and Y. Tagawa, ``Electron spectroscopy studies on magneto-optical media and plastic substrate interface,'' IEEE Transl. J. Magn. Japan, vol. 2, pp. 740--741, August 1987 [Digests 9th Annual Conf. Magnetics Japan, p. 301, 1982].
% \bibitem{b7} M. Young, The Technical Writer's Handbook. Mill Valley, CA: University Science, 1989.
% \end{thebibliography}
% \vspace{12pt}

\end{document}